\pdfoutput=1
\documentclass[11pt]{article}

\usepackage[]{coling}

\usepackage{times}
\usepackage{latexsym}

\usepackage[T1]{fontenc}
\usepackage[utf8]{inputenc}

\usepackage{microtype}
\usepackage{hyperref}
\usepackage{url}
\usepackage{graphicx}
\usepackage{booktabs}
\usepackage{subcaption}
\usepackage{booktabs}
\usepackage{cleveref}
\usepackage{placeins}
\usepackage{amssymb}
\usepackage{enumitem}

\title{How Many Languages Make Good Multilingual Instruction Tuning? \\ A Case Study on BLOOM}

\author{Shaoxiong Ji\thanks{Equal contribution.}\thanks{Work done while at the University of Helsinki.} \\
  Technical University of Darmstadt \\ University of Helsinki \\
  \texttt{shaoxiong.ji@tu-darmstadt.de} \\\And
  Pinzhen Chen\footnotemark[1] \\
  University of Edinburgh \\
  \texttt{pinzhen.chen@ed.ac.uk} \\}

\begin{document}
\maketitle
\begin{abstract}
Instruction tuning a large language model with multiple languages can prepare it for multilingual downstream tasks. Nonetheless, it is yet to be determined whether having a handful of languages is sufficient, or whether the benefits increase with the inclusion of more. By fine-tuning large multilingual models on 1 to 52 languages, we present a case study on BLOOM to understand three pertinent factors affecting performance: the number of languages, language exposure, and similarity between training and test languages. Overall we found that 1) expanding language coverage in multilingual instruction tuning proves to be beneficial; 2) accuracy often significantly boots if the test language appears in the instruction mixture; 3) languages' genetic features correlate with cross-lingual transfer more than merely the number of language but different languages benefit to various degrees.

\end{abstract}

\section{Introduction}

Many large language models (LLMs) have been designed to handle many languages through multilingual pre-training, like mGPT~\citep{shliazhko2022mgpt} and BLOOM~\citep{scao2022bloom}, while some other models only (officially) support a few, e.g.~the Llama series~\citep{touvron2023llama,grattafiori2024llama3herdmodels}. 
At the alignment stage, as a more affordable route, researchers used multilingual instruction tuning (mIT) to enhance the multilingualism of LLM \citep{muennighoff-etal-2023-crosslingual}. Recently, \citet{chen-etal-2024-monolingual} compared monolingual and multilingual instruction tuning under a resource-fair scenario with multiple LLMs. \citet{kew2023turning} experimented with English-centric LLMs such as Llama 2~\citep{touvron2023llama} and Falcon~\citep{almazrouei2023falcon} and found that mIT using as few as three languages enables cross-lingual transfer. 
Similarly, \citet{shaham2024multilingual} studied the same topic featuring ``just a pinch of multilinguality'' of 2--4 languages. 
Moreover, \citet{chirkova2024zero} showed that instruction tuning in only English with a carefully set learning rate enables responses in four other test languages. We note the variations in the choice of languages, base models, and testbeds used in these studies. 
More importantly, while it has been demonstrated that a handful of languages elicit (zero-shot) multilingual responses, it does not imply the optimal downstream task results, not to mention to cater to each language. 

To fill the gap, we perform instruction tuning on the multilingual BLOOM model~\citep{scao2022bloom} on a parallel instruction dataset named Bactrain-X in 52 languages~\citep{li2023bactrianx}. 
We progressively add a language for each mIT run, resulting in 52 models in total,\footnote{\href{https://huggingface.co/collections/MaLA-LM/lucky52-660e5fd24a2ced4b334d63d6}{huggingface.co/collections/MaLA-LM $\rightarrow$~Lucky52}. Fun fact: \href{https://en.wikipedia.org/wiki/Lucky_52}{Lucky 52} was a famous variety show in China in the early 2000s. The naming denotes the 52 models fine-tuned on 1 to 52 languages.} which are then evaluated on three multilingual benchmarks. 
Patterns on BLOOM reveal that \textit{contrary to prior research, having more languages beyond a handful can further improve performance, although with diminishing returns and some outlier cases.} 
Our findings are summarized as follows:
\begin{enumerate}
    \item Cross-lingual transfer improves with more languages in mIT, but the optimal number of languages depends on the task and test language, with varying behaviours across benchmarks and languages.
    \item Including a specific language in the instruction tuning data generally enhances its performance, though outliers exist, and the benefits from massive mIT are limited if a language is not part of the tuning data, regardless of its presence during pre-training.
    \item Correlations between language similarity and performance vary, with genetic features being more predictive than the number of languages. Some languages, like Thai and Swahili, show strong inter-language effects, while others, like English and Chinese, have weaker correlations.
\end{enumerate}

Our study emphasizes the importance of a closer look at the tasks, benchmarks, languages, and evaluation metrics. 
We advocate for more consistent future studies focused on mIT. Further variables include but are not limited to base LLMs, pre-training recipes and data. Comprehensive and consistent investigations are crucial for advancing our understanding of mIT and its implications.

\section{Scaling Instruction Languages}

\subsection{Increasing the Number of Languages}

Our setup is supervised fine-tuning, where an instruction and a task input are fed to an LLM to yield a response. 
We progressively include an extra language in each training run to study the precise effect the number of languages brings in. 
As the number of languages expands, instruction data size also grows---to mitigate this variable, we opt for parallel instruction data in which English instructions are translated into other languages. 
This controls that all models are trained with a comparable amount of instruction information. 
Moreover, the increase in data size also increases the number of optimization steps when utilizing stochastic gradient descent to update the model parameters on the same device. 
We express the number of updates as $U = \lceil\frac{N\times L \times E}{B\times W}\rceil$,
where $N$ is the instruction data size, $L$ is the number of languages, $E$ is the number of epochs, $B$ is the batch size, and $W$ is the number of GPUs. 
We increase $W$ proportionally to $L$ to maintain a manageable range of updates.

\subsection{Multilingual Instruction Data}

We use the Bactrian-X dataset \citep{li2023bactrianx} comprising 3.4 million instruction-response pairs with an equal share in each of its 52 languages. 
We fine-tune an LLM from 1 to 52 languages resulting in 52 models. The languages are added in a specific order: en (English), zh (Chinese), and the rest in alphabetical order. Refer to \Cref{app:languages} for an exhaustive list of languages and their 2-digit codes.

\subsection{Base Language Model Tuning}

Multilingual language models can inherit distribution biases present in the training data, which may affect their capability across languages after instruction tuning.
We hence base our experiments on BLOOM~\citep{scao2022bloom} which has been developed with careful consideration in multiple natural and coding languages. 
Its massive multilingual tokenization support makes it well-suited for studying a large number of languages in a fair manner.
In our specific application, we opt for the BLOOM-7B1 variant with 7.1B parameters. 
\paragraph{Training Details}

We use the transformers framework~\citep{wolf2019huggingface} with DeepSpeed integration~\citep{rasley2020deepspeed} for fine-tuning.
We set the learning rate to 3e-5 and the batch size to 4 per device.
Gradient accumulation, with a step size of 4, enables the aggregation of gradients over multiple steps.
The number of epochs is fixed at 3.
The maximum model length is set to 768, the same as in Bactrian-X~\citep{li2023bactrianx}.
Models are trained on a cluster with 4 AMD MI250X GPUs (in total 8 GPU dies) in each node. 
We adopt distributed training on multiple nodes ranging from 2 to 10 with the increase in the number of languages, making the global batch size range from 256 to 1280.

\begin{figure}[t]
    \centering
    \includegraphics[width=0.7\linewidth,trim=0ex 2ex 0ex 4ex,clip]{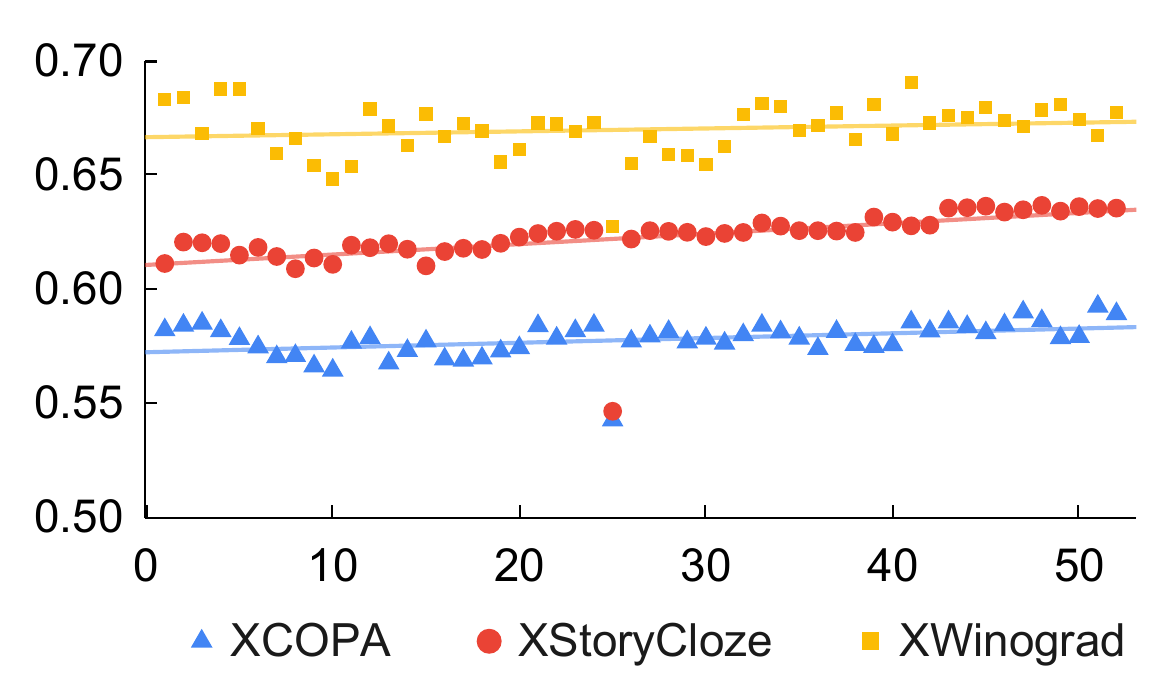}
    \vspace{-1ex}
    \caption{Test performance across all languages; x-axis: number of languages in mIT; y-axis: average accuracy.}
    \label{fig:average-accuracy}
\end{figure}

\subsection{Benchmarks and Evaluation}
We test on three multilingual benchmarks. 
XCOPA~\citep{ponti-etal-2020-xcopa} is a multilingual dataset for causal commonsense reasoning in 11 languages. 
XStoryCloze~\citep{lin2021few} is a multilingual dataset for commonsense reasoning in the story in 10 non-English languages. 
XWinograd~\citep{tikhonov-ryabinin-2021-heads} is a multilingual compilation of Winograd Schemas~\citep{levesque2012winograd} available in 6 languages. 
We run zero-shot prompting via lm-evaluation-harness~\citep{eval-harness}. 
Different models, trained with progressively added languages, are evaluated on these benchmarks using accuracy (0-1) as the metric.

\begin{figure*}[t]
    \centering
    \begin{minipage}{0.32\linewidth}
        \centering
        \includegraphics[width=\linewidth,trim=0ex 7ex 0ex 4ex,clip]{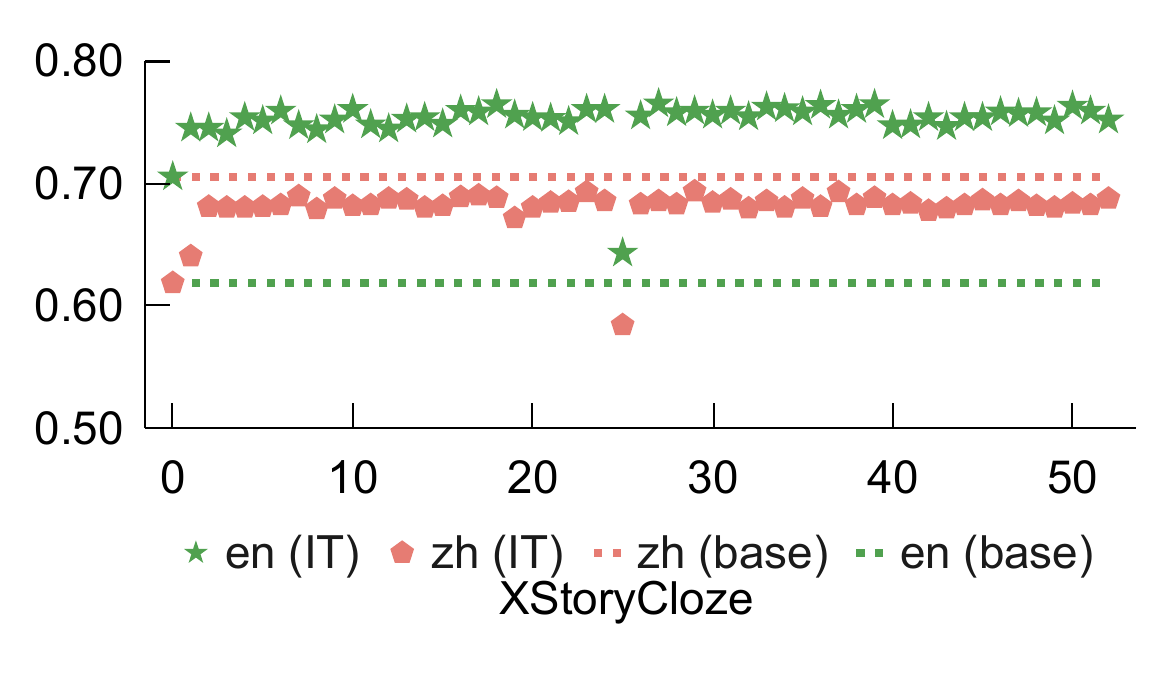}
    \end{minipage}%
    \hfill
    \begin{minipage}{0.32\linewidth}
        \centering
        \includegraphics[width=\linewidth,trim=0ex 7ex 0ex 4ex,clip]{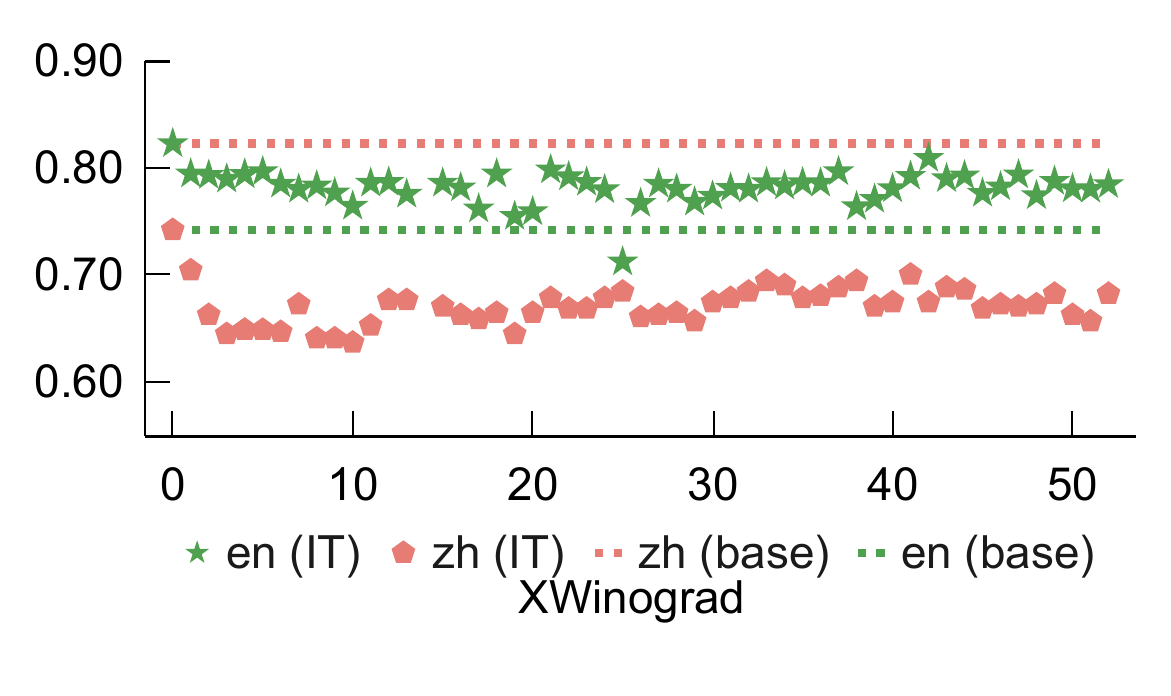}
    \end{minipage}%
    \hfill
    \begin{minipage}{0.32\linewidth}
        \centering
        \includegraphics[width=\linewidth,trim=0ex 7ex 0ex 4ex,clip]{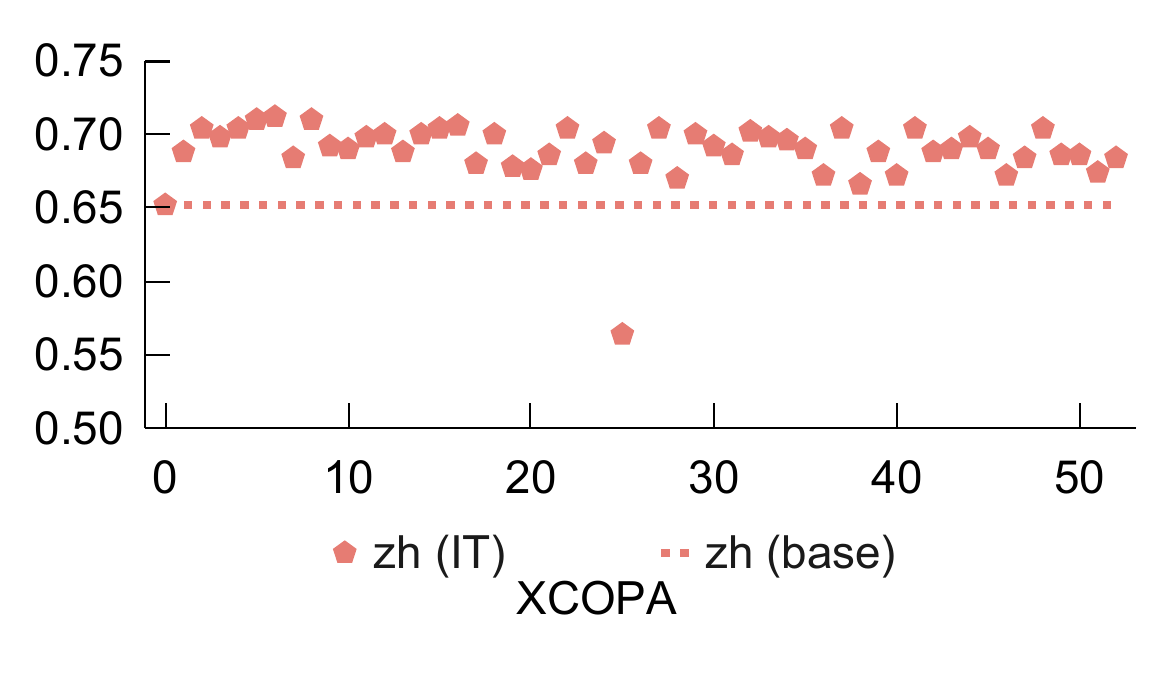}
    \end{minipage}
    \vspace{-1ex}
    \caption{Accuracy for English and Chinese on XStoryCloze, XWinograd, and XCOPA.}
    \label{fig:zh-en-two-benchmarks}
\end{figure*}

\begin{figure}[t]
\centering
    \includegraphics[width=0.6\linewidth,trim=0ex 6ex 0ex 2ex,clip]{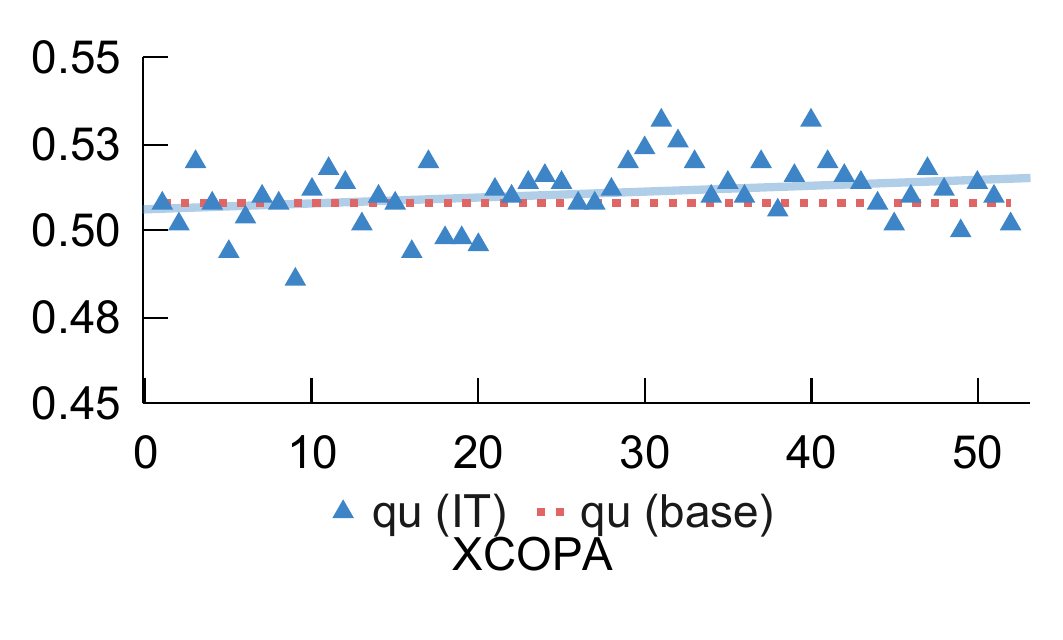}
    \vspace{-1ex}
    \caption{Accuracy for Quechuan, unseen both by the base model and during IT.}
    \label{fig:base-unseen-instruct-unseen-xcopa-xwinograd}
\end{figure}

\begin{figure*}[t]
    \centering
    \begin{minipage}{0.235\textwidth}
        \centering
        \includegraphics[width=\linewidth,trim=0ex 10ex 0ex 4ex,clip]{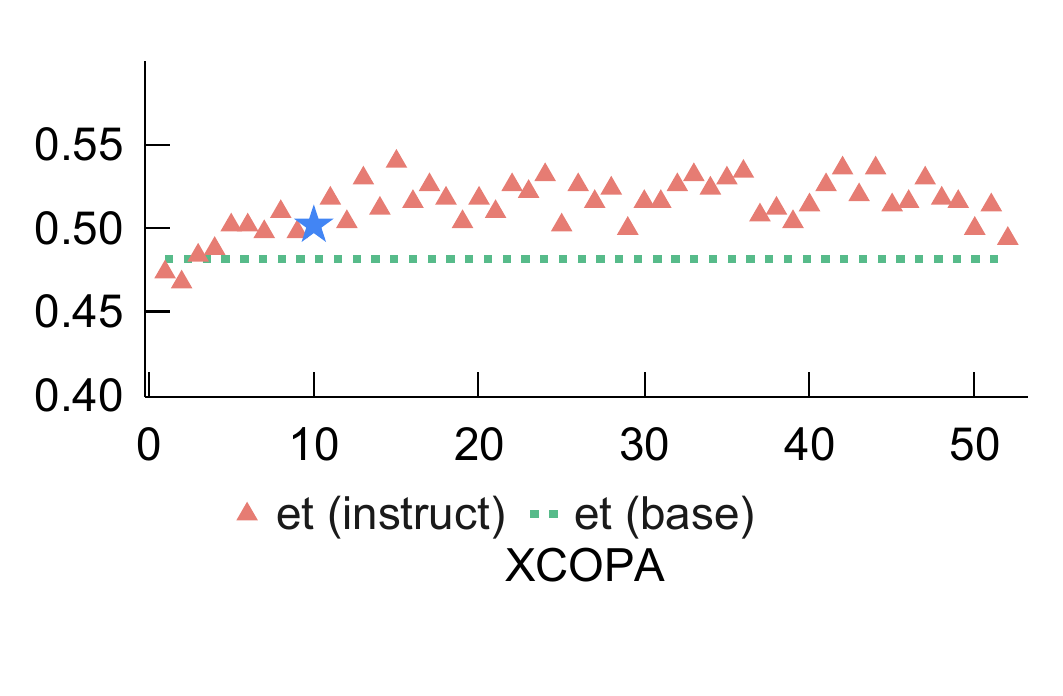}
        \label{fig:adding-is-useful-xcopa-et}
    \end{minipage}
    \hfill
    \begin{minipage}{0.235\textwidth}
        \centering
        \includegraphics[width=\linewidth,trim=0ex 10ex 0ex 4ex,clip]{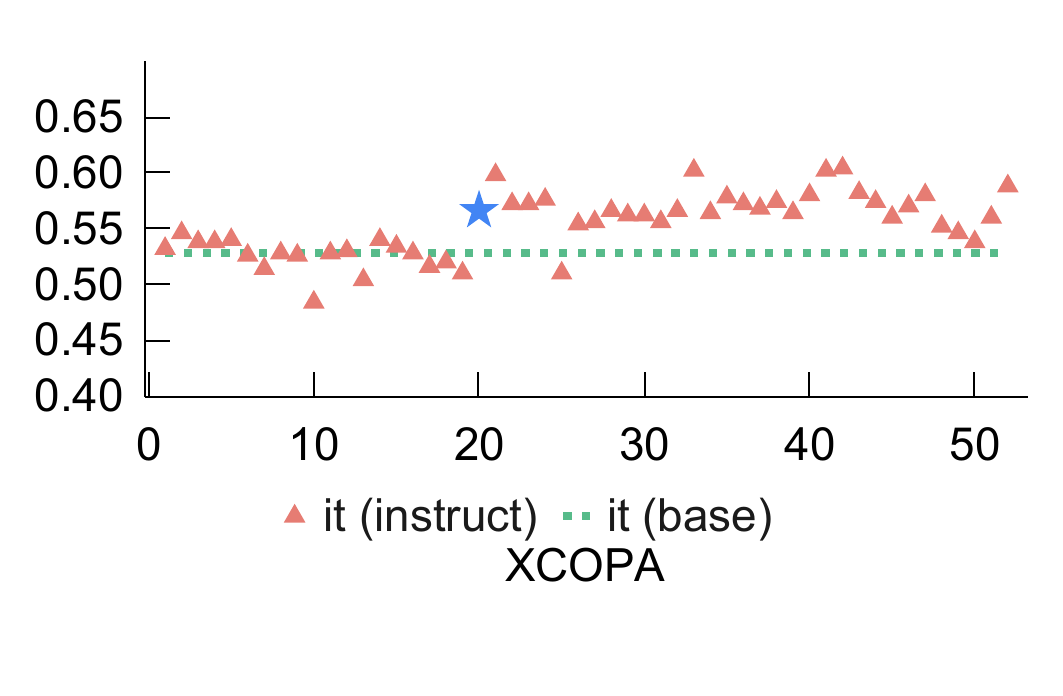}
    \label{fig:adding-is-useful-xcopa-it}
    \end{minipage}
    \hfill
    \begin{minipage}{0.235\textwidth}
        \centering
        \includegraphics[width=\linewidth,trim=0ex 10ex 0ex 4ex,clip]{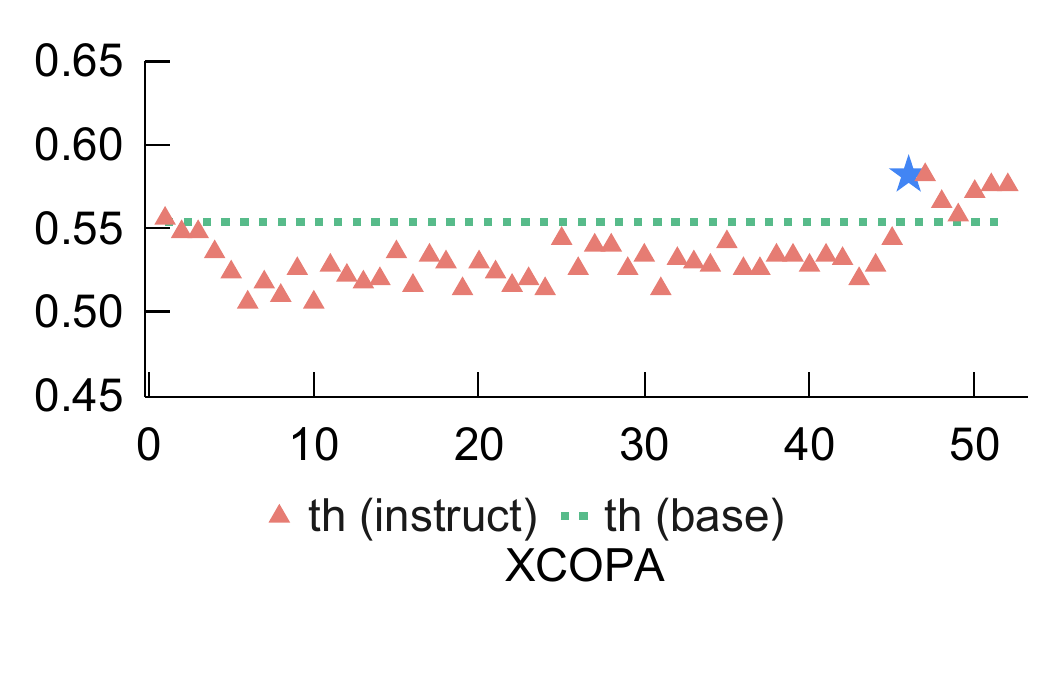}
    \label{fig:adding-is-useful-xcopa-th}
    \end{minipage} 
    \hfill
    \begin{minipage}{0.235\textwidth}
    \centering
    \includegraphics[width=\linewidth,trim=0ex 10ex 0ex 2ex,clip]{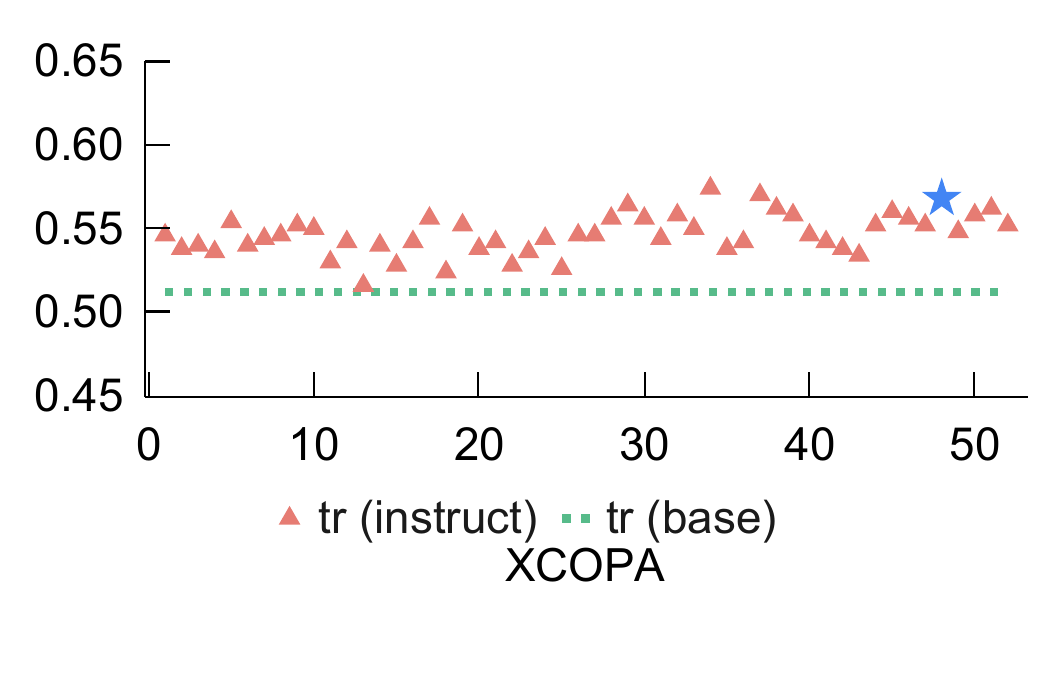}
    \label{fig:adding-is-useless-xcopa-th}
    \end{minipage}
    \vspace{-4ex}
    \caption{Accuracy on XCOPA for various languages, unseen by the base model but seen during IT. \textcolor{blue!75}{\footnotesize$\bigstar$} indicates the point the test language starts to be included in the mIT data. In most cases, performance can benefit (et, it, th) from the test language appearing in mIT despite outliers (tr).}
    \label{fig:base-unseen-instruct-seen-xcopa}
\end{figure*}

\section{Results and Discussions}
\subsection{Number of Languages}

\paragraph{Overall pattern} We first study the effect of the number of languages on multilingual performance---how much multilingualism do we want for instruction tuning an LLM, i.e., BLOOM-7B1 in our case study?
\Cref{fig:average-accuracy} illustrates the average accuracy on the three benchmarks with different numbers of languages in the instruction data. 
For XCOPA and XStoryCloze, there is a positive correlation between the number of instruction languages and performance; but for XWinograd, we observe fluctuating results with a weaker trend. A notable drop appears in the scatter plot across all benchmarks---the point where Korean (kr) is added to the IT data. 
We inspected the training curve of the model trained with Korean added to instruction data and found that the training and validation loss decreases as training goes on and the model converges as expected.

\paragraph{English versus Chinese} We move on to two specific languages, i.e. English and Chinese, as displayed in \Cref{fig:zh-en-two-benchmarks} together with base model prompting performance. We notice a similar drop in accuracy when Korean is added, but there is no obvious benefit from cross-lingual transfer when more languages are added. 
For English and Chinese XStoryCloze, the highest accuracy is attained much later when the 27th (Latvian, lv) or 29th (Malaysian, ml) language is added, respectively.
Yet, interestingly, while instruction tuning surpasses the base model for English, it makes it worse for Chinese.  
XWinograd exhibits a similar trend that instruction tuning benefits English but drastically harms Chinese. In addition, the best IT performance for both languages is observed when there is only one language (English) in the instruction data. 
Specifically, the result for Chinese XCOPA peaks early when the 6th language (Bangla, bn) is added, but later models with more languages no longer improve.

\paragraph{Summary} Instruction tuning with a few languages is useful for cross-lingual transfer, but having more languages can further improve the average results when many languages are of concern. 
However, distinct behaviours can be witnessed for different benchmarks and individual languages, so the optimal number of languages in mIT depends on the task and test language.

\subsection{Language Exposure}

A test language can fall into one of the cases depending on being seen or unseen during the pre-training and instruction tuning phases: (1) unseen by the base and during IT: qu (XCOPA). (2) seen by the base but unseen during IT: ht (XCOPA) and eu (XStoryCloze). (3) unseen by the base but seen during IT: et, it, th, tr (XCOPA); my, ru (XStoryCloze); ja, ru (XWinograd). (4) seen by the base and during IT, e.g.: id, sw, ta, vi, zh (XCOPA); ar, en, es, hi, id, sw, te, zh (XStoryCloze); en, fr, pt, zh (Winograd). We are interested in understanding model performance in the first three categories where the number and closeness of mIT languages may benefit or harm unseen languages.

\paragraph{Unseen by base, unseen during IT}

Only one language is not covered by either pre-training or instruction tuning: Quechuan (qu) in XCOPA. We plot its performance across all data mixtures in \Cref{fig:base-unseen-instruct-unseen-xcopa-xwinograd}. 
As the number of IT languages grows, accuracy fluctuates around the base model performance, showing a weak trend. This implies that if a language has no presence at all, there is very little transfer mIT can do.

\paragraph{Unseen by base, seen during IT}
We then investigate an important use of multilingual instruction tuning---to adapt the base LLM to unseen languages during pre-training. 
\Cref{fig:base-unseen-instruct-seen-xcopa} show the accuracy of various languages that have not been (intentionally) learned during pre-training but appeared in instruction tuning (at some point), with additional languages exhibiting similar trends in \Cref{app:lang-exposure} \Cref{fig:base-unseen-instruct-seen}. 
We find that in the majority of scenarios, including a language in IT can immediately aid the performance of that language as anticipated.
However, we also notice two interesting cases: 1) for Turkish (tr) tested in XCOPA (\Cref{fig:base-unseen-instruct-seen-xcopa}), the accuracy is similar before and after introducing the language to the IT data; 2) for Russian (ru) tested in XWinograd (\Cref{fig:adding-is-useless}), the performance is below base model prompting even after the language appears in mIT. 
Besides, cross-lingual transfer is observed in mIT, for instance: the performance of Estonia (et) and Italian (it) tested in XCOPA can further grow after more languages are added; the performance of Turkish (tr) tested in XCOPA is already favourable without the language itself. 

\begin{figure}[t]
\centering
    \begin{minipage}{0.235\textwidth}
    \centering
    \includegraphics[width=\linewidth,trim=0ex 7ex 0ex 2ex,clip]{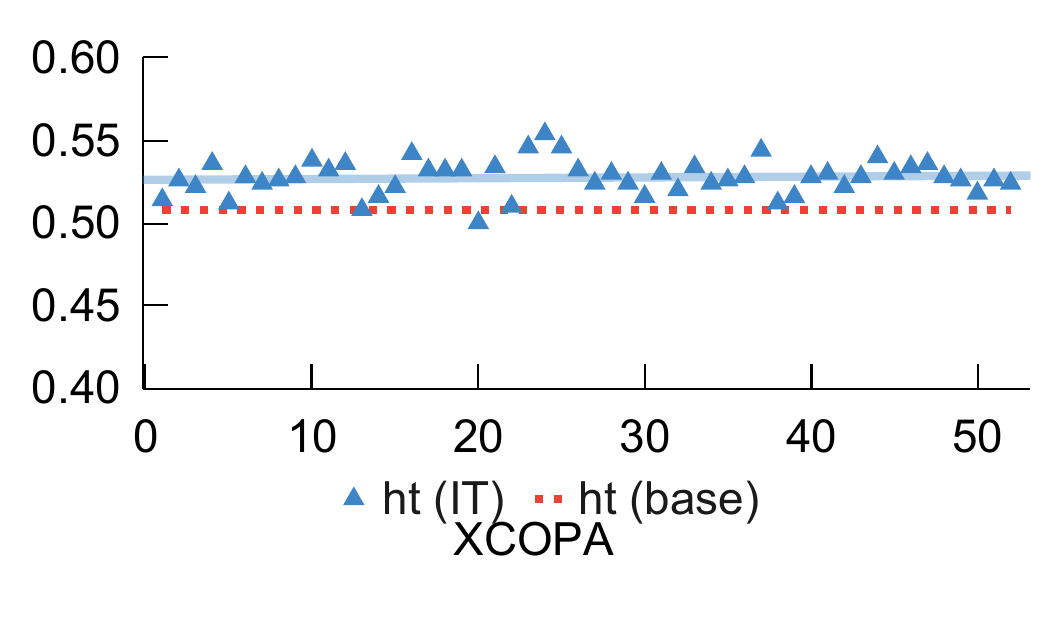}
    \end{minipage}
    \hfill
    \begin{minipage}{0.235\textwidth}
    \centering
    \includegraphics[width=\linewidth,trim=0ex 7ex 0ex 2ex,clip]{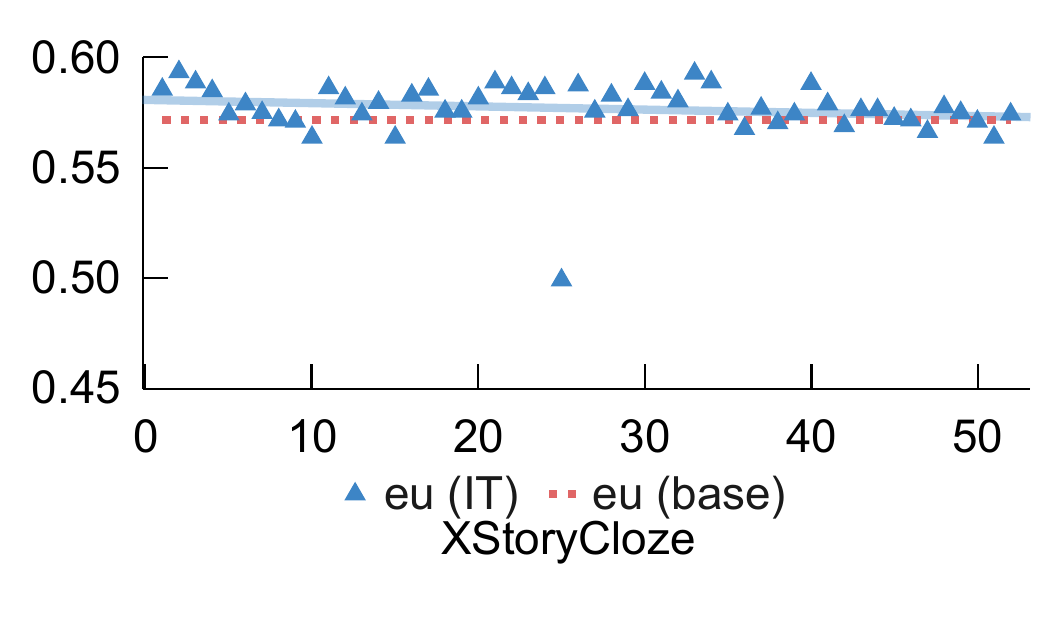}
    \end{minipage}
    \vspace{-1ex}
    \caption{Accuracy for Haitian on XCOPA and Basque on XStoryCloze, seen by base unseen during IT.}
    \label{fig:base-seen-instruct-unseen-xcopa-xstory}
\end{figure}

\paragraph{Seen by base, unseen during IT}
Finally, \Cref{fig:base-seen-instruct-unseen-xcopa-xstory} displays the two languages in this category. 
In both cases, IT is better than base model prompting, indicating that mIT can transfer to unseen IT languages that have been pre-trained. Nonetheless, increasing the number of languages during mIT does not significantly bring benefits.

\paragraph{Summary} By examining result patterns in exhaustive seen-unseen cases, we can infer that having a particular language in instruction tuning data is often beneficial for its performance, although some outliers can be observed. Regardless of whether a language is learned during pre-training, if it does not appear in the IT data composite, the benefit from massive mIT is usually limited.

\subsection{Language Similarity}

We conduct a post-hoc analysis on how language closeness affects cross-lingual transfer. Instead of studying the relation between the number of fine-tuning languages and test set performance, we define an aggregated similarity measure between all languages present in a fine-tuning corpus and a test language $L_{\mathrm{test}}$:

$\mathrm{similarity}_{\mathrm{train},\mathrm{test}} = \sum_{L\in \mathrm{corpus}} \mathrm{sim}(L, L_\mathrm{test})$

\noindent where $\mathrm{sim}(,)$ is a similarity metric between two languages. We measure ``aggregated similarity'' instead of ``average similarity'' because we argue that, given their giant sizes, LLMs have the capacity to model all language data in the training set simultaneously.

We adopt different similarity measures based on syntactic, geographic, phonological, genetic, inventory, and featural distances scored by \texttt{lang2vec} \citep{littell-etal-2017-uriel,malaviya-etal-2017-learning}.\footnote{\href{https://github.com/antonisa/lang2vec}{github.com/antonisa/lang2vec}} 
In addition, we gathered from another source a language closeness score derived from sound (consonants) overlap, which is deemed to reflect genetic similarity  \citep{beaufils2020stochastic}.\footnote{\href{http://www.elinguistics.net/language_evolution.html}{elinguistics.net/language\_evolution.html}}
In total, we test out seven measures, where the similarity score is always normalized to between 0 and 1 to the lowest and highest similarity. The choice of language features is similar to a contemporaneous study on language transferability and similarity \citep{philippy-etal-2024-soft}. 
As a baseline comparison, we provide Pearson correlation coefficients between the number of languages and performance: XStoryClose in \Cref{tab:xstorycloze-correlation}, XWingrad in \Cref{tab:xwinograd-correlation}, and XCOPA in \Cref{tab:xcopa-correlation}. Also, since empirically the addition of Korean leads to an outlying performance, we compute coefficients without the particular checkpoint too.

\begin{table*}[thb]
\centering\footnotesize
\begin{tabular}{lcccccccccc}
\toprule
& \textbf{ar} & \textbf{en} & \textbf{es} & \textbf{hi} & \textbf{id} & \textbf{my} & \textbf{ru} & \textbf{sw} & \textbf{te} & \textbf{zh} \\
\midrule
num. lang. & -0.07 & 0.15 & 0.46 & 0.51 & 0.53 & 0.75 & 0.81 & 0.56 & -0.47 & 0.11 \\
num. lang. w/o ko & \phantom{-}0.08 & \textbf{0.41} & \textbf{0.73} & \textbf{0.66} & 0.63 & 0.75 & 0.86 & 0.56 & \textbf{-0.53} & 0.31 \\
\midrule
sound correspondence & -0.06 & 0.15 & 0.48 & 0.52 & 0.57 & 0.82 & 0.83 & 0.67 & -0.43 & 0.12 \\
lang2vec featural & -0.05 & 0.15 & 0.47 & 0.51 & 0.53 & 0.77 & 0.83 & 0.58 & -0.46 & 0.13 \\
lang2vec genetic  & \phantom{-}0.17 & 0.16 & 0.50 & 0.54 & \textbf{0.66} & \textbf{0.96} & \textbf{0.87} & \textbf{0.96} & -0.26 & \textbf{0.37} \\
lang2vec geographic  & \phantom{-}0.17 & 0.15 & 0.47 & 0.51 & 0.54 & 0.76 & 0.81 & \textbf{0.96} & -0.48 & \textbf{0.37} \\
lang2vec inventory & -0.06 & 0.15 & 0.46 & 0.51 & 0.54 & 0.76 & 0.83 & 0.55 & -0.46 & 0.13 \\
lang2vec phonological & -0.05 & 0.15 & 0.47 & 0.51 & 0.53 & 0.76 & 0.83 & 0.57 & -0.45 & 0.13 \\
lang2vec syntactic & -0.05 & 0.15 & 0.47 & 0.51 & 0.53 & 0.78 & 0.82 & 0.57 & -0.45 & 0.13 \\
\bottomrule   
\end{tabular}
\vspace{-1ex}
\caption{Pearson correlation between XStoryCloze performance and mIT data similarity}\label{tab:xstorycloze-correlation}
\end{table*}

\begin{table*}[thb]
\centering\footnotesize
\begin{tabular}{lcccccc}
\toprule
& \textbf{en} & \textbf{fr} & \textbf{ja} & \textbf{pt} & \textbf{ru} & \textbf{zh} \\
\midrule
num. lang. & -0.02 & 0.01 & 0.62 & -0.32 & -0.07 & 0.49 \\
num. lang. w/o ko & -0.03 & 0.00 & 0.66 & -0.35 & -0.07 & \textbf{0.50} \\
\midrule
sound correspondence & -0.01 & -0.01\phantom{-} & 0.66 & -0.33 & -0.06 & 0.45 \\
lang2vec featural & -0.01 & 0.00 & 0.62 & -0.31 & -0.06 & 0.47 \\
lang2vec genetic & -0.02 & -0.08\phantom{-} & \textbf{0.72} & -0.35 & -0.05 & -0.31\phantom{-}       \\
lang2vec geographic & -0.02 & -0.01\phantom{-} & 0.62 & -0.33 & -0.07 & -0.31\phantom{-}       \\
lang2vec inventory & -0.01 & 0.00 & 0.62 & -0.31 & -0.06 & 0.48 \\
lang2vec phonological & -0.01 & 0.01 & 0.63 & -0.32 & -0.06 & 0.48 \\
lang2vec syntactic & -0.02 & 0.00 & 0.62 & -0.32 & -0.06 & 0.47 \\
\bottomrule    
\end{tabular}
\vspace{-1ex}
\caption{Pearson correlation between XWinograd performance and mIT data similarity}\label{tab:xwinograd-correlation}
\end{table*}

\begin{table}[ht]
\centering\scriptsize
\setlength{\tabcolsep}{0.45ex}
\begin{tabular}{lccccccccc}
\toprule
 & \textbf{et} &  \textbf{id} &  \textbf{it} &  \textbf{sw} &  \textbf{ta} &  \textbf{th} &  \textbf{tr} &  \textbf{vi} &  \textbf{zh} \\
\midrule
\footnotesize num. lang. & 0.44 & 0.44 & 0.63 & 0.54 & -0.80 & 0.53 & 0.45 & -0.46 & -0.20 \\
\footnotesize num. lang. w/o ko & 0.44 & 0.50 & 0.64 & 0.54 & -0.80 & 0.53 & 0.46 & \textbf{-0.50} & \textbf{-0.39} \\
\midrule
\footnotesize sound correspond. & 0.51 & 0.48 & 0.64 & 0.64 & -0.83 & 0.62 & 0.45 & -0.36 & -0.20 \\
\footnotesize l2v featural & 0.46 & 0.45 & 0.63 & 0.56 & -0.81 & 0.55 & 0.45 & -0.44 & -0.19 \\
\footnotesize l2v genetic & \textbf{0.67} & \textbf{0.58} & \textbf{0.67} & \textbf{0.93} & \textbf{-0.84} & \textbf{0.82} & \textbf{0.47} & \phantom{-}0.02 & \phantom{-}0.01 \\
\footnotesize l2v geographic & 0.43 & 0.46 & 0.64 & \textbf{0.93} & -0.80 & 0.55 & 0.45 & -0.45 & \phantom{-}0.01 \\
\footnotesize l2v inventory & 0.46 & 0.45 & 0.64 & 0.52 & -0.80 & 0.55 & 0.45 & -0.45 & -0.19 \\
\footnotesize l2v phonological & 0.45 & 0.45 & 0.62 & 0.54 & -0.80 & 0.55 & 0.44 & -0.45 & -0.19 \\
\footnotesize l2v syntactic & 0.45 & 0.45 & 0.63 & 0.54 & -0.81 & 0.54 & 0.45 & -0.45 & -0.19 \\
\bottomrule
\end{tabular}
\vspace{-1ex}
\caption{Pearson correlation between XCOPA performance and mIT data similarity}\label{tab:xcopa-correlation}
\end{table}

For XCOPA and XStoryCloze, \texttt{lang2vec} genetic features stand out, usually resulting in a stronger correlation than simply the number of languages. While most languages display a positive correlation with mIT language similarity or coverage, we notice that some are negatively affected: ta, and vi in XCOPA, te in XStoryCloze, and pt in XWinograd. Finally, across different test sets, behaviours could be diverging for the same language: genetic similarity benefits ru in XStoryCloze but has no correlation in XWinograd.

\paragraph{Summary} Many factors contribute to the train-test similarity and performance correlation in both positive and negative ways: languages, test sets, and similarity measures.

\FloatBarrier
\section{Conclusion}

While instruction tuning of large multilingual models enables versatile language processing, it requires careful handling of language-specific nuances. This paper presents an experimental analysis that controls the base model, instructions, and training recipe to study the number, closeness, and exposure of languages. Our findings, compared with prior work, show that multilingual instruction tuning depends heavily on factors like base models, data, tasks, and evaluation protocols. We emphasize the need for more systematic studies to validate the effectiveness and generalizability of this approach.

\section*{Limitations}
Our work studies multilingual instruction tuning in 52 relatively high-resourced languages, which might be limited in size to arrive at comprehensive conclusions for thousands of living languages, which are often under-served. 
We did not conduct a human evaluation due to budget constraints. 
Future work could conduct a more systematic assessment with more rigorously controlled variables and heavier regularization during instruction tuning to prevent base model knowledge and language forgetting.

\section*{Acknowledgments}
This work has received funding from the European Union's Horizon Europe research and innovation programme under grant agreement No 101070350 and from UK Research and Innovation (UKRI) under the UK government’s Horizon Europe funding guarantee [grant number 10052546]. 

We acknowledge CSC-IT Center for Science, Finland for awarding this project access to the LUMI supercomputer, owned by the EuroHPC Joint Undertaking, hosted by CSC (Finland) and the LUMI consortium through Finnish extreme scale call (project LumiNMT) and Czech Republic allocations issued by e-INFRA CZ, and IT4Innovations National Supercomputing Center. 

\bibliography{multilinguality}

\appendix
\begin{figure*}
\section{All languages}\label{app:languages}
Apart from English and Chinese, data in the other 50 languages in Bactrian-X are added in alphabetical order: 
af (Afrikaans), ar (Arabic), az (Azerbaijani), bn (Bengali), cs (Czech), de (German), es (Spanish), et (Estonian), fa (Farsi), fi (Finnish), fr (French), gl (Galician), gu (Gujarati), he (Hebrew), hi (Hindi), hr (Croatian), id (Indonesian), it (Italian), ja (Japanese), ka (Georgian), kk (Kazakh), km (Khmer), ko (Korean), lt (Lithuanian), lv (Latvian), mk (Macedonian), ml (Malayalam), mn (Mongolian), mr (Marathi), my (Burmese), ne (Nepali), nl (Dutch), pl (Polish), ps (Pashto), pt (Portuguese), ro (Romanian), ru (Russian), si (Sinhala), sl (Slovenian), sv (Swedish), sw (Swahili), ta (Tamil), te (Telugu), th (Thai), tl (Tagalog), tr (Turkish), uk (Ukrainian), ur (Urdu), vi (Vietnamese), and xh (Xhosa).
\end{figure*}

\begin{figure*}[t]
\section{Additional plots for languages unseen by base model but seen during IT}
\label{app:lang-exposure}
\vspace{2ex}
    \centering
    \begin{subfigure}{\textwidth}
    \begin{minipage}{0.32\textwidth}
        \centering
        \includegraphics[width=\linewidth,trim=0ex 0ex 0ex 0ex,clip]{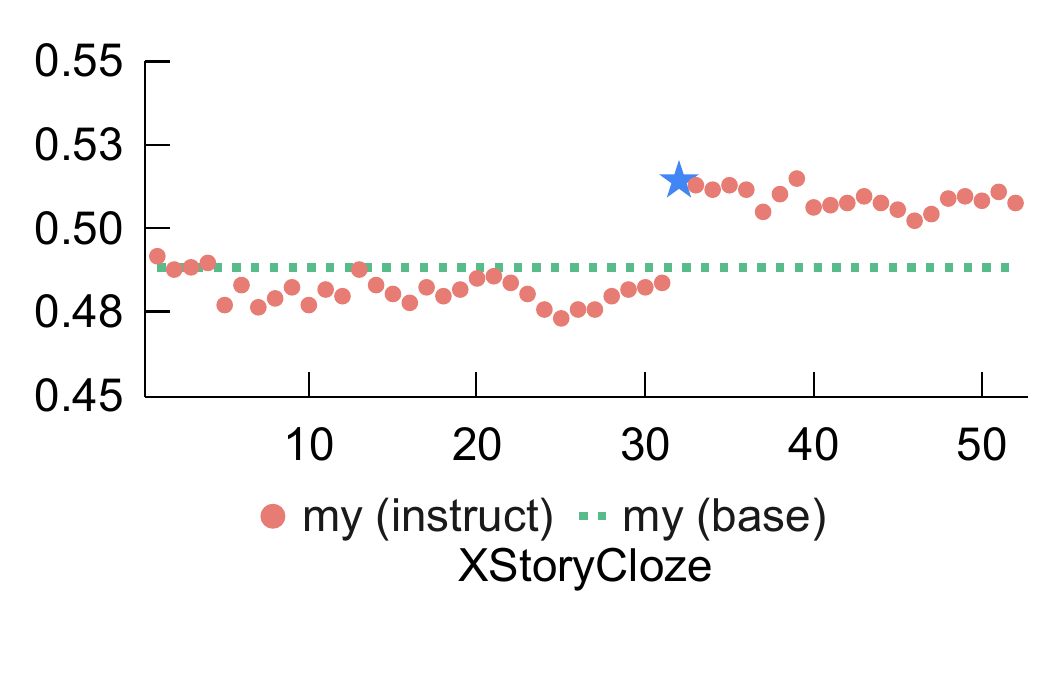}
    \end{minipage}
    \hfill
    \begin{minipage}{0.32\textwidth}
        \centering
        \includegraphics[width=\linewidth,trim=0ex 0ex 0ex 0ex,clip]{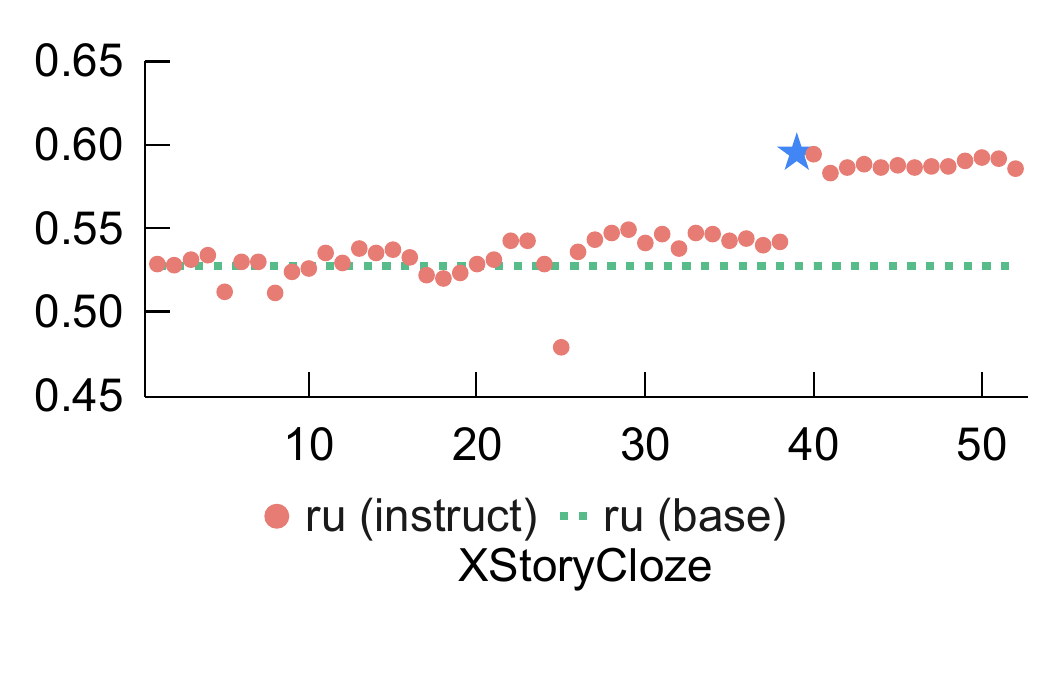}
    \end{minipage}
    \hfill
    \begin{minipage}{0.32\textwidth}
        \centering
        \includegraphics[width=\linewidth,trim=0ex 0ex 0ex 0ex,clip]{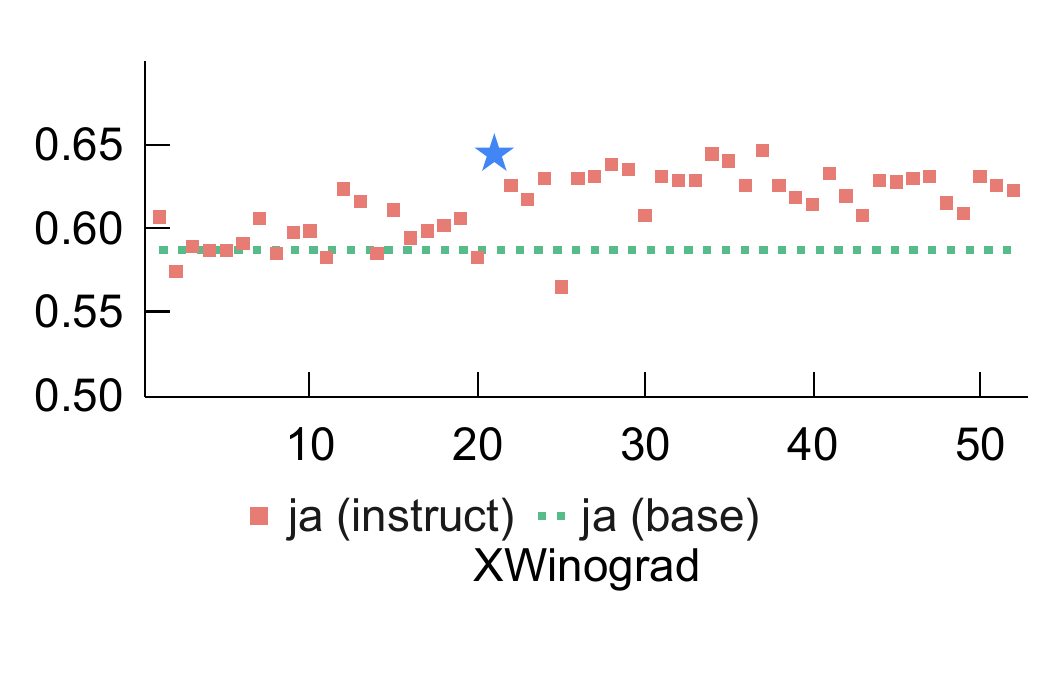}
    \end{minipage}
    \caption{Performance on XStory Malaysian and Russian as well as XWinograd Japanese benefits from the test language changing from unseen to seen in mIT.}
    \label{fig:adding-is-useful}
    \end{subfigure}

    \begin{subfigure}{\textwidth}
    \centering
    \begin{minipage}{0.32\textwidth}
        \centering
        \includegraphics[width=\linewidth,trim=0ex 0ex 0ex 0ex,clip]{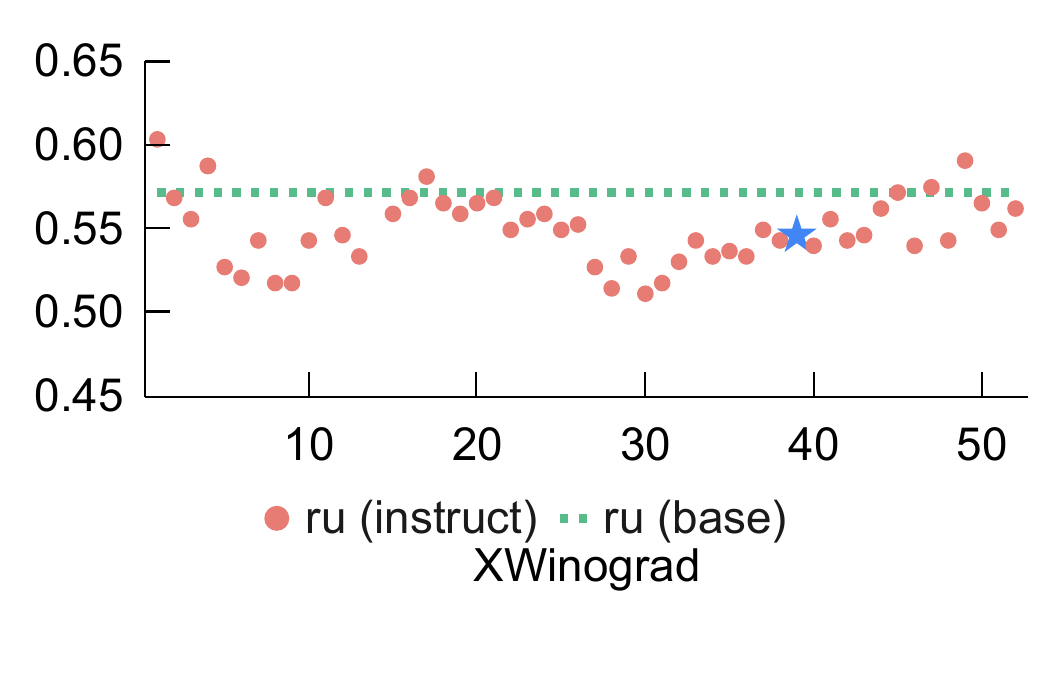}
    \end{minipage}
    \caption{Performance on XWinograd Russian does not benefit from the test language changing from unseen to seen in mIT.}
    \label{fig:adding-is-useless}
    \end{subfigure}
    \caption{Accuracy for various languages unseen by the base model but seen during IT. \textcolor{blue!75}{\footnotesize$\bigstar$} indicates the point when the test language starts to be included in the mIT data.}
    \label{fig:base-unseen-instruct-seen}
\end{figure*}

\end{document}